\documentclass{article}
\usepackage{ijcai22}

\usepackage{times}
\usepackage{soul}
\usepackage{url}
\usepackage[hidelinks]{hyperref}
\usepackage[utf8]{inputenc}
\usepackage[small]{caption}
\usepackage{graphicx}
\usepackage{amsmath}
\usepackage{amsthm}
\usepackage{booktabs}
\usepackage{algorithm}
\usepackage{algorithmic}
\usepackage{subfigure}
\usepackage{multirow} 
\usepackage{booktabs} 
\usepackage[normalem]{ulem} 
\usepackage[dvipsnames]{xcolor}
\definecolor{mygray}{gray}{0.6}

\title{Federated clustering with GAN-based data synthesis}

\author{
Jie Yan
\and
Jing Liu
\and
Ji Qi
\And
Zhong-Yuan Zhang$^\star$
\affiliations
School of Statistics and Mathematics, \\ Central University of Finance and Economics, Beijing, P.R.China
\emails
zhyuanzh@gmail.com
}

\begin{document}

\maketitle

\begin{abstract}
Federated clustering (FC) is an extension of centralized clustering in federated settings. The key here is how to construct a global similarity measure without
sharing private data, since the local similarity may be insufficient to group local data correctly and the similarity of samples across clients cannot be directly measured due to privacy constraints. Obviously, the most straightforward way to analyze FC is to employ the methods extended from centralized ones, such as K-means (KM) and fuzzy c-means (FCM). However, they are vulnerable to non independent-and-identically-distributed (non-IID) data among clients. To handle this, we propose a new federated clustering framework, named synthetic data aided federated clustering (SDA-FC). It trains generative adversarial network locally in each client and uploads the generated synthetic data to the server, where KM or FCM is performed on the synthetic data. The synthetic data can make the model immune to the non-IID problem and enable us to capture the global similarity characteristics more effectively without sharing private data. Comprehensive experiments reveals the advantages of SDA-FC, including superior performance in addressing the non-IID problem and the device failures.

\end{abstract}

\section{Introduction}
In an era where privacy concerns are paramount, federated learning \cite{mcmahan2017communication} has garnered considerable attention and found applications in diverse domains, including autonomous driving \cite{nguyen2022deep,fu2023incentive}, smart healthcare \cite{nguyen2022federated}, smart cities \cite{rasha2023federated} and IoT-data \cite{nguyen2021federated,he2023clustered}. It aims to train a global model by fusing multiple local models trained on client devices without sharing private data.

\begin{figure}[!t]
\centering
\subfigure[Client 1]{
\includegraphics[width=1.5in]{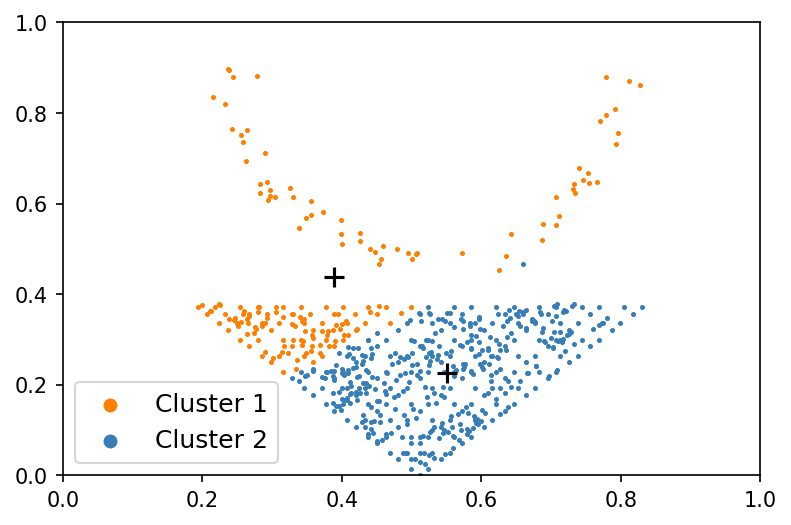}}
\quad
\subfigure[Client 2]{
\includegraphics[width=1.5in]{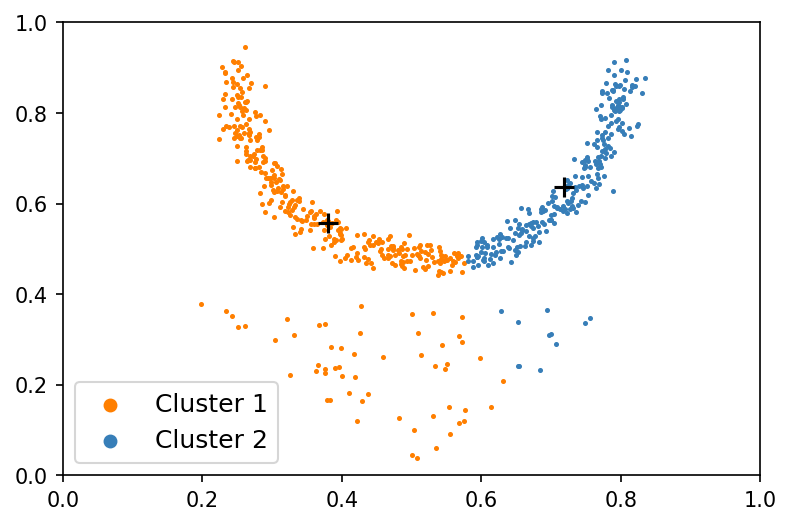}}

\subfigure[Global real dataset]{
\includegraphics[width=1.5in]{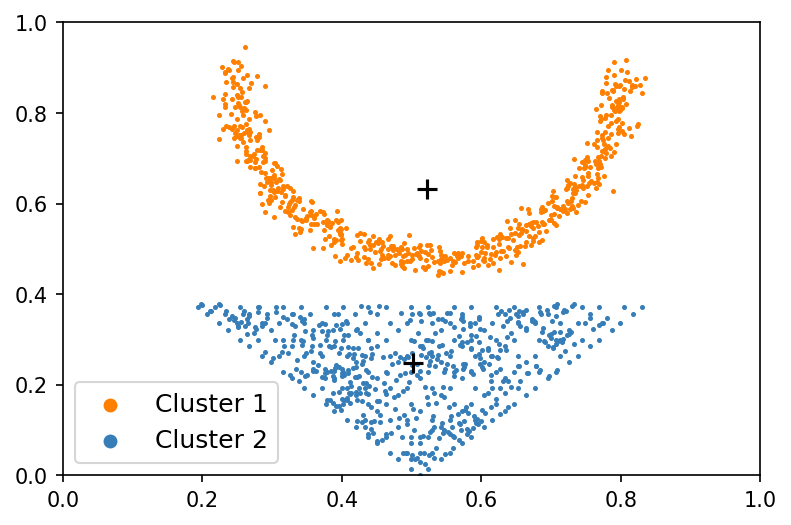}}
\quad
\subfigure[Global synthetic dataset]{
\includegraphics[width=1.5in]{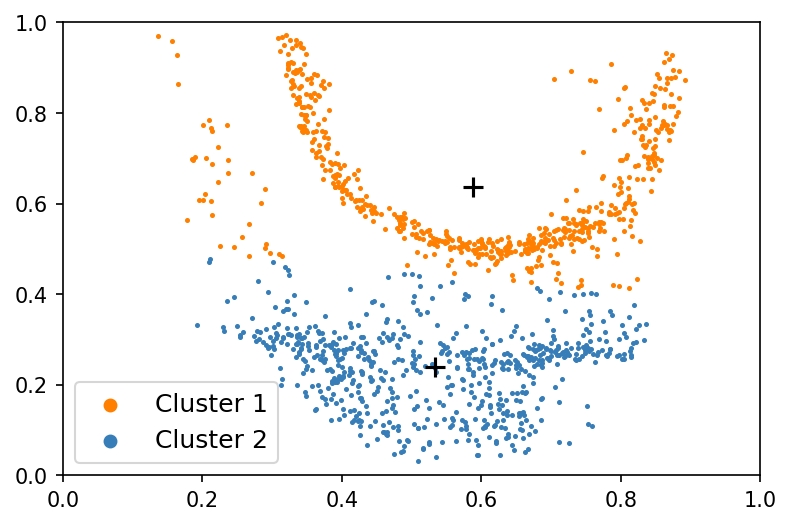}}
\caption{A toy example to illustrate the inherent interest of federated clustering. (a)-(b) The local data on two client devices. (c) All local data are combined in a central storage. (d) An approximation of the global real dataset. For each dataset, we perfromed  K-means on it. Cluster centroids are indicated by "+" and samples are colored by the clustering results. Only (c) correctly identifies two clusters, and the centroids in (d) are closest to those in (c).}
\label{case_1_vis}
\end{figure}

\begin{table}[!t]
\caption{NMI of clustering methods on the toy example. The results indicate that there is a gap between state-of-the-art federated clustering algorithms and classical centralized clustering algorithms, but it is narrowed by us.}
\renewcommand{\arraystretch}{1.25} 
\tabcolsep 1.625mm 
\begin{tabular}{cccccc}
\hline\hline
\multicolumn{3}{c}{KM-based methods} &\multicolumn{3}{c}{FCM-based methods}\\ \cmidrule(r){1-3} \cmidrule(r){4-6}
\textcolor{mygray}{KM\_central} &k-FED &ours & \textcolor{mygray}{FCM\_central} &FFCM &ours\\
\hline
1 &0.7352 &1  & 1 &0.3691 &1\\

\hline\hline
\end{tabular}\label{case_1_nmi}
\end{table}

While federated learning excels in independent-and-identically-distributed (IID) scenarios, the local data distributions across client devices often deviate substantially from the IID scenarios. This phenomenon, referred to as non-IID data or data heterogeneity, can impede convergence and detrimentally affect model performance \cite{zhu2021federated,ma2022state}. In non-IID scenarios, a natural approach involves departing from the conventional single-center framework, which trains only one global model. Instead, one can construct a multi-center framework, harnessing the power of client clustering \cite{ghosh2020efficient,zec2022decentralized} or data clustering to enhance collaboration \cite{dennis2021heterogeneity,stallmann2022towards,chung2022federated}. For the client clustering, the basic idea of it is that each client may come from a specific distribution. Thus, we should use clients within the same cluster to collaboratively train a specific global model. However, each sample within a single client may also come from a specific distribution. Hence, the data clustering, also know as \textbf{federated clustering}, can be more beneficial for client collaboration \cite{stallmann2022towards}. The goal of it is to cluster data based on a global similarity measure while keeping all data local \cite{stallmann2022towards}.

Beyond its role in mitigating the non-IID problem, federated clustering itself presents an intriguing research avenue. As shown in Fig. \ref{case_1_vis}, local similarity alone inadequately recovers accurate local data grouping, while a global perspective excels in this regard. Yet, obtaining the global real dataset is unfeasible due to the confidentiality of local client data. Hence, the key here is how to measure the global similarity without sharing private data. To handle this, previous work introduced adaptations of classic centralized clustering algorithms, such as K-means (KM) \cite{macqueen1967classification} and fuzzy c-means (FCM) \cite{bezdek1984fcm}, to the federated setting, resulting in k-FED \cite{dennis2021heterogeneity} and federated fuzzy c-means (FFCM) \cite{stallmann2022towards}. The basic ideas of them are similar: alternately estimating the local cluster centroids and the global ones, in other words, mining the local centroids based on local private data and uploading them to the server, where one can run KM to construct $k$ global cluster centroids for housing the global similarity information that is then disseminated to clients. However, the resulting global cluster centroids may be delicate and susceptible to varying non-IID levels, thereby impairing the robustness and performance of the model.

Given a federated dataset, the non-IID level solely hinges on the local data distribution rather than the global one. Consequently, the model performance can exhibit insensitivity to different non-IID levels if a good approximation of the global data can be constructed. Moreover, such an approximation can enhance the capture of global similarity characteristics without compromising data privacy. Motivated by this, we propose a simple but effective federated clustering framework, denominated \textbf{synthetic data aided federated clustering (SDA-FC)}. It encompasses two main steps: global synthetic data construction and cluster assignment. In the first step, the central server aims to construct a global synthetic data using multiple local GANs \cite{goodfellow2014generative} trained from local data. In the second step, the central server first performs KM or FCM on the global synthetic dataset to get $k$ global cluster centroids that is then disseminated to clients. Subsequently, cluster assignment is performed based on the proximity of local data to these centroids. As shown in Fig. \ref{case_1_vis}, the local cluster centroids (Fig. \ref{case_1_vis} (a) and Fig. \ref{case_1_vis} (b)) markedly differ from the global ones (Fig. \ref{case_1_vis}. (c)), and the latter cannot be accurately approximated by running KM on the former simply. Notably, the global synthetic data (Fig. \ref{case_1_vis} (d)) emerges as a good approximation of the global real data, with centroids derived from this synthetic dataset closely mirroring those obtained from the real global data. To further sense the gap between federated clustering and centralized clustering, we performed k-FED and FFCM on the toy example with the federated setting and performed KM and FCM with the centralized setting. As shown in Table \ref{case_1_nmi}, there is a gap between them, but it is narrowed by us.

In addition to the non-IID problem, there are also some other challenges in federated learning, such as expensive communication and systems heterogeneity \cite{li2020federated}. These challenges can lead to low efficiency and device failures. Nevertheless, the proposed framework effectively addresses these concerns, requiring only one communication round between the central server and clients, and can run asynchronously while remaining robust to device failures.

The rest of this paper is organized as follows: Sect.\ref{related_work} first concisely describe four core challenges in federated learning, and then review some representative federated clustering methods. Following that, Sect. \ref{sect3} first introduce some preliminaries about GAN, and then propose a new federated clustering framework SDA-FC. Sect. \ref{sect4} reveals the advantages of SDA-FC. Finally, this paper is concluded in Sect. \ref{sect6}.

\section{Related Work}\label{related_work}
\subsection{Core challenges in federated learning}
In traditional centralized learning, all data is typically stored on a central server. For this learning protocol, one problem is hardware constraints that it places a high demand on the storage and computing capacity of the central server when the dataset is large. And another problem is privacy risks and data leakage \cite{zhu2021federated}. To handle the first problem, distributed data parallelism \cite{dean2012large} uses multiple devices to train model replicas in parallel using different data subsets. Nevertheless, this method still necessitates access to the entire dataset for division into subsets, thus retaining inherent privacy risks.

Federated learning \cite{mcmahan2017communication}, a recent research focal point, offers a promising solution by enabling the collaborative training of a global model through the fusion of local models trained on client devices, all while safeguarding user privacy. Despite its potential to alleviate both hardware and privacy concerns, federated learning presents several challenges, which are outlined as follows \cite{li2020federated}:
\begin{itemize}
\item \textbf{Expensive Communication}.
In practice, federated networks may encompass millions of client devices, rendering network communication substantially more costly than local computation \cite{huang2013depth,van2009multi}. Crafting communication-efficient models involves minimizing both communication rounds and message volumes.

\item \textbf{Systems Heterogeneity}.
Client devices often possess diverse communication, computation, and storage capabilities due to variations in wireless network connectivity, energy resources, and hardware limitations. Consequently, only a fraction of devices may effectively participate in training a large model \cite{bonawitz2019towards}, and device disconnections during training must be anticipated. Thus, a robust federated model should accommodate low device participation and device failures.

\item \textbf{Data Heterogeneity}.
In federated learning, local data distributions across client devices can differ substantially, influenced by how each client utilizes their local device. This phenomenon, known as non-IID data or data heterogeneity, can impede convergence and degrade model performance \cite{zhu2021federated}. A natural solution to this challenge involves moving beyond the traditional single-center framework, where only one global model is trained. Instead, constructing a multi-center framework, which simultaneously trains multiple global models based on client clustering \cite{ghosh2020efficient,zec2022decentralized} or data clustering \cite{dennis2021heterogeneity,stallmann2022towards,chung2022federated}, may offer a more viable approach.

\item  \textbf{Privacy Concerns}.
Although the classic federated averaging method \cite{mcmahan2017communication} helps mitigate shallow data leakage from raw data, Zhu et al. \cite{zhu2019deep} demonstrated its potential to lead to deep data leakage through shared gradients and parameters. This deep data leakage allows the reconstruction of original data from these gradients and parameters. To enhance privacy in federated learning, various works have employed differential privacy \cite{dwork2006calibrating,dwork2011firm,dwork2014algorithmic}, a popular privacy protection technique \cite{geyer2017differentially,zhao2020local,wei2020federated}. However, while improving privacy, differential privacy can significantly degrade model performance.
\end{itemize}

Numerous centralized algorithms have been adapted for federated settings in recent years, encompassing reinforcement learning \cite{wang2019edge,wang2020optimizing,zhang2021deep}, classification algorithms \cite{brisimi2018federated,yu2020federated,chen2021bridging}, clustering algorithms \cite{dennis2021heterogeneity,stallmann2022towards,chung2022federated}, and more. Nevertheless, these extensions are often discouraged by the challenges of federated learning in terms of model efficiency and model performance. This work specifically addresses the challenges associated with federated clustering.

\subsection{Federated clustering}
As shown in Fig. \ref{case_1_vis}, the local similarity is insufficient to group local data correctly, while a global perspective excels in this regard. However, the global real dataset cannot be obtained since the local data on client devices are forbidden to share. Hence, the key here is how to measure the global similarity without sharing private data.

To tackle this challenge, k-FED \cite{dennis2021heterogeneity} and federated fuzzy c-means (FFCM) \cite{stallmann2022towards} have extended two classic centralized clustering algorithms, K-means (KM) \cite{macqueen1967classification} and fuzzy c-means (FCM) \cite{bezdek1984fcm}, to the federated setting, respectively. Specifically, each client device runs a classic centralized clustering algorithm on the local data to generate several cluster centroids and uploads them to the central server. Then, the central server can construct $k$ global cluster centroids by running KM on the uploaded local cluster centroids. The classic centralized clustering algorithm used in k-FED is KM and that used in FFCM is FCM. However, the resulting global cluster centroids often exhibit fragility and sensitivity to varying non-IID levels, leading to suboptimal and non-robust model performance.

Given a federated dataset, the non-IID level quantifies the heterogeneity degree among local data distributions, and it remains independent of the global distribution. Hence, model performance can remain unaffected by different non-IID levels if a good approximation of the global real data can be constructed. Additionally, such an approximation may facilitate the more effective capture of global similarity characteristics without sharing private data. Motivated by this, we introduce a simple but effective federated clustering framework based on GAN-based data synthesis, named synthetic data aided federated clustering (SDA-FC). It requires only one communication round, can run asynchronously, and can handle device failures. It comprises two main steps: global synthetic data construction and cluster assignment. In the former, the central server endeavors to construct a global synthetic dataset using multiple local GANs \cite{goodfellow2014generative}, trained on local data. For the latter, the central server first applies KM or FCM to the global synthetic dataset, resulting in $k$ global cluster centroids. Then, the final clustering result can be obtained based on the proximity of local data points to these centroids.

\section{Synthetic Data Aided Federated Clustering (SDA-FC) }
\label{sect3}
In this section, we present the synthetic data aided federated clustering (SDA-FC) framework. First, we introduce some preliminaries about GAN, which form the basis of our approach. Subsequently, we detail the SDA-FC.

\subsection{Preliminaries}
The vanilla GAN \cite{goodfellow2014generative} architecture comprises two networks: the generator and the discriminator. These networks engage in a two-player game wherein they continuously improve each other. The generator aims to produce synthetic samples that closely resemble real data, thereby deceiving the discriminator, whose role is to differentiate between real and generated samples. The game concludes when the discriminator can no longer distinguish between real and generated samples, indicating that the generator has learned the underlying real data distribution and reached the theoretical global optimum.

The objective function of the vanilla GAN is defined as:
\begin{equation}
\mathop{\min}\limits_{G}\mathop{\max}\limits_{D}\mathop{\mathbf{E}}
\limits_{z \sim \mathcal{N}}\log(1 - D(G(z))) + \mathop{\mathbf{E}} \limits_{x \sim p_r}\log(D(x)),
\end{equation}
where $G$ is the generator that inputs a noise $z$ and outputs a generated sample, $\mathcal{N}$ is Gaussian distribution, $D$ is the discriminator that inputs a sample and outputs a scalar to tell the generated samples from the real ones, and $p_r$ is the distribution of real data. While GAN have demonstrated remarkable success in various applications, adversarial training is known for its instability and susceptibility to mode collapses \cite{Luke2017}, resulting in both high-quality and low-diversity generated samples. Mode collapses imply that the model captures only a portion of the real data's characteristics. To address this, Mukherjee et al. \cite{mukherjee2019clustergan} introduced an additional categorical variable into the generator's input, promoting a clearer cluster structure in the latent space and enhancing sample diversity.

To mitigate mode collapses in our SDA-FC framework, we also use a mixture of discrete and continuous variables as the input of the generator by following \cite{mukherjee2019clustergan}. The new objective function is defined as:
\begin{equation}
\mathop{\min}\limits_{G}\mathop{\max} \limits_{D}\mathop{\mathbf{E}}
\limits_{u \sim \mathcal{U}} \mathop{\mathbf{E}}\limits_{z \sim \mathcal{N}}\log(1 - D(G(e_u,\, z))) + \mathop{\mathbf{E}} \limits_{x \sim p_r}\log(D(x)),
\label{GAN}
\end{equation}
where $\mathcal{U}$ is a uniform random distribution with the lowest value 1 and the highest value $k$, and $e_u$ is a one-hot vector with the $u$-th element being 1.

\subsection{Synthetic Data Aided Federated Clustering (SDA-FC)}
Given a real world dataset $X$, which is distributed among $m$ clients, i.e. $X=\bigcup_{i=1}^{m} X^{(i)}$. The goal of federated clustering is to cluster data based on a global similarity measure while keeping all data local \cite{stallmann2022towards}. However, direct measurement of global similarity is infeasible without sharing private data. This is why federated clustering is a challenging task to handle.

To handle this, we propose the synthetic data aided federated clustering (SDA-FC) framework, which is designed to require only a single round of communication between clients and the central server. It consists of two main steps: global synthetic data construction and cluster assignment, which are detailed below.

\subsubsection{Global synthetic data construction}
Firstly, client $i$ ($i = 1,\, 2,\, \cdots, \, m$) downloads an initial GAN model from the central server and trains it with the local data $X^{(i)}$. Then, the trained generator $G^{(i)}$ and the local data size of $X^{(i)}$ are uploaded to the central server. Finally, the central server use $G^{(i)}$ to generate a dataset $\hat{X}^{(i)}$ of the same size as $X^{(i)}$, and the global synthetic dataset $\hat{X}$ can be obtained by merging all generated datasets, i.e. $\hat{X}=\bigcup_{i=1}^{m} \hat{X}^{(i)}$.

\subsubsection{Cluster assignment}
As a general federated clustering framework, it is simple to combine SDA-FC with some centralized clustering methods, such as K-means (KM) \cite{macqueen1967classification} and fuzzy c-means (FCM) \cite{bezdek1984fcm}, leading to two specific methods, SDA-FC-KM and SDA-FC-FCM. Specifically, the central server first performs KM or FCM on the global synthetic dataset to obtain $k$ global centroids. Then, each client downloads the centroids and the final clustering result can be obtained by performing cluster assignment according to the cosine distance from local data to the centroids.

\section{Experimental results}
\label{sect4}
In this section, we detail the experimental settings, validate the efficacy of our proposed methods and the global synthetic dataset on several datasets with different non-IID scenarios, and perform a sensitivity analysis of clustering methods in response to device failures induced by system heterogeneity. Finally, we summarize the experimental findings.

\begin{table}[!t]
\caption{Description of datasets.}
\scalebox{0.8}{
\renewcommand{\arraystretch}{1.5} 
\tabcolsep 2mm 
\begin{tabular}{ccccc}
\hline\hline
\textbf{Dataset} & \textbf{Type} &\textbf{Size} & \textbf{Image size/Features} & \textbf{Class} \\\hline
MNIST           & image        & 70000    & $28\times28$   & 10\\
Fashion-MNIST   & image        & 70000    & $28\times28$   & 10\\
CIFAR-10        & image        & 60000    & $32\times32$            & 10 \\
STL-10          & image        & 13000    & $96\times96$           & 10 \\
Pendigits       & time series  & 10992    & 16             & 10\\

\hline\hline
\label{datasets}
\end{tabular}}
\end{table}

\subsection{Experimental Settings}
Creating a universal non-IID benchmark dataset in the realm of federated learning remains a challenging endeavor due to the inherent complexity of federated learning itself \cite{li2022federated,hu2022oarf}. Here, following ref. \cite{chung2022federated}, we simulate different federated scenarios by partitioning the real-world dataset into $k$ smaller subsets (each one corresponds to a client) and scaling \textbf{the non-IID level} $p$ in clients, where $k$ is the number of true clusters. For a client with $s$ datapoints, the first $p\cdot s$ datapoints are sampled from a single cluster, and the remaining $(1 - p)\cdot s$ ones are randomly sampled from any cluster. Extremely, $p = 0$ means the local data across the clients is IID, and $p = 1$ means the local data across the clients is completely non-IID.

As shown in Table \ref{datasets}, we select two grey image datasets MNIST and Fashion-MNIST, two color image datasets CIFAR-10 \cite{krizhevsky2009learning} and STL-10 \footnote{Note that, to reduce the computational cost of baseline methods and to use the same network structure for CIFAR-10, we performed a preprocessing step to resize the images in STL-10 to 32 $\times$ 32.} \cite{coates2011analysis}, and a time series dataset Pendigits \cite{keller2012hics} for comprehensive analysis. In SDA-FC, all local GANs are trained with the Adam Optimizer \cite{kingma2014adam}.  More detailed hyperparameter settings can be found in
the Appendix A. Our code will be made publicly available to facilitate reproducibility.

\begin{table}[!t]
\caption{NMI of clustering methods in different federated scenarios. For each comparison, the best result is highlighted in boldface.}
\scalebox{0.59}{
\renewcommand{\arraystretch}{1.6} 
\tabcolsep 1.3mm 
\begin{tabular}{cccccccc}
\hline\hline
\multirow{2}{*}{Dataset} &\multirow{2}{*}{$p$} &\multicolumn{3}{c}{KM-based methods} &\multicolumn{3}{c}{FCM-based methods}\\ \cmidrule(r){3-5} \cmidrule(r){6-8}
\quad &\quad &\textcolor{mygray}{KM\_central} &k-FED &SDA-FC-KM &\textcolor{mygray}{FCM\_central} &FFCM &SDA-FC-FCM \\

\hline
\multirow{5}{*}{MNIST} &0.0 &\multirow{5}{*}{\textcolor{mygray}{0.5304}} &0.5081 &\textbf{0.5133} &\multirow{5}{*}{\textcolor{mygray}{0.5187}} &\textbf{0.5157} &0.5141\\
\quad &0.25 &\quad &0.4879 &\textbf{0.5033} &\quad &\textbf{0.5264} &0.5063\\
\quad &0.5 &\quad &0.4515 &\textbf{0.5118} &\quad &0.4693 &\textbf{0.5055}\\
\quad &0.75 &\quad &0.4552 &\textbf{0.5196} &\quad &0.4855 &\textbf{0.5143}\\
\quad &1.0 &\quad &0.4142 &\textbf{0.5273} &\quad &\textbf{0.5372} &0.5140\\

\hline
\multirow{5}{*}{Fashion-MNIST} &0.0 &\multirow{5}{*}{\textcolor{mygray}{0.6070}} &0.5932 &\textbf{0.5947} &\multirow{5}{*}{\textcolor{mygray}{0.6026}} &0.5786 &\textbf{0.6027}\\
\quad &0.25 &\quad &0.5730 &\textbf{0.6052} &\quad &\textbf{0.5995} &0.5664\\
\quad &0.5 &\quad &\textbf{0.6143} &0.6063 &\quad &\textbf{0.6173} &0.6022\\
\quad &0.75 &\quad &0.5237 &\textbf{0.6077} &\quad &\textbf{0.6139} &0.5791\\
\quad &1.0 &\quad &0.5452 &\textbf{0.6065} &\quad &0.5855 &\textbf{0.6026}\\

\hline
\multirow{5}{*}{CIFAR-10} &0.0 &\multirow{5}{*}{\textcolor{mygray}{0.0871}} &0.0820 &\textbf{0.0823} &\multirow{5}{*}{\textcolor{mygray}{0.0823}} &0.0812 &\textbf{0.0819}\\
\quad &0.25 &\quad &\textbf{0.0866} &0.0835 &\quad &\textbf{0.0832} &0.0818\\
\quad &0.5 &\quad &\textbf{0.0885} &0.0838 &\quad &\textbf{0.0870} &0.0810\\
\quad &0.75 &\quad &0.0818 &\textbf{0.0864} &\quad &\textbf{0.0842} &0.0808\\
\quad &1.0 &\quad &\textbf{0.0881} &0.0856 &\quad &0.0832 &\textbf{0.0858}\\

\hline
\multirow{5}{*}{STL-10} &0.0 &\multirow{5}{*}{\textcolor{mygray}{0.1532}} &0.1468 &\textbf{0.1470} &\multirow{5}{*}{\textcolor{mygray}{0.1469}} &\textbf{0.1436} &0.1406\\
\quad &0.25 &\quad &0.1472 &\textbf{0.1511} &\quad &\textbf{0.1493} &0.1435\\
\quad &0.5 &\quad &0.1495 &\textbf{0.1498} &\quad &0.1334 &\textbf{0.1424}\\
\quad &0.75 &\quad &\textbf{0.1455} &0.1441 &\quad &0.1304 &\textbf{0.1425}\\
\quad &1.0 &\quad &0.1403 &\textbf{0.1477} &\quad &\textbf{0.1565} &0.1447\\

\hline
\multirow{5}{*}{Pendigits} &0.0 &\multirow{5}{*}{\textcolor{mygray}{0.6877}} &\textbf{0.7001} &0.6972 &\multirow{5}{*}{\textcolor{mygray}{0.6862}} &\textbf{0.6866} &0.6654\\
\quad &0.25 &\quad &0.6620 &\textbf{0.6796} &\quad &\textbf{0.6848} &0.6631\\
\quad &0.5 &\quad &0.6625 &\textbf{0.6661} &\quad &\textbf{0.6798} &0.6544\\
\quad &0.75 &\quad &0.5521 &\textbf{0.6734} &\quad &0.6757 &\textbf{0.6846}\\
\quad &1.0 &\quad &\textbf{0.6296} &0.5172 &\quad &\textbf{0.7236} &0.5162\\

\hline
count &- &- &7 &18 &- &16 &9\\

\hline\hline
\end{tabular}\label{NMI}}
\end{table}

\begin{table}[!t]
\caption{Kappa of clustering methods in different federated scenarios. For each comparison, the best result is highlighted in boldface.}
\scalebox{0.59}{
\renewcommand{\arraystretch}{1.6} 
\tabcolsep 1.3mm 
\begin{tabular}{cccccccc}
\hline\hline
\multirow{2}{*}{Dataset} &\multirow{2}{*}{$p$} &\multicolumn{3}{c}{KM-based methods} &\multicolumn{3}{c}{FCM-based methods}\\ \cmidrule(r){3-5} \cmidrule(r){6-8}
\quad &\quad &\textcolor{mygray}{KM\_central} &k-FED &SDA-FC-KM &\textcolor{mygray}{FCM\_central} &FFCM &SDA-FC-FCM \\

\hline
\multirow{5}{*}{MNIST} &0.0 &\multirow{5}{*}{\textcolor{mygray}{0.4786}} &\textbf{0.5026} &0.4977 &\multirow{5}{*}{\textcolor{mygray}{0.5024}} &0.5060 &\textbf{0.5109}\\
\quad &0.25 &\quad &0.4000 &\textbf{0.4781} &\quad &\textbf{0.5105} &0.5027\\
\quad &0.5 &\quad &0.3636 &\textbf{0.4884} &\quad &0.3972 &\textbf{0.4967}\\
\quad &0.75 &\quad &0.3558 &\textbf{0.4926} &\quad &0.4543 &\textbf{0.5021}\\
\quad &1.0 &\quad &0.3386 &\textbf{0.5000} &\quad &\textbf{0.5103} &0.5060\\

\hline
\multirow{5}{*}{Fashion-MNIST} &0.0 &\multirow{5}{*}{\textcolor{mygray}{0.4778}} &\textbf{0.4657} &0.4640 &\multirow{5}{*}{\textcolor{mygray}{0.5212}} &0.4974 &\textbf{0.5252}\\
\quad &0.25 &\quad &\textbf{0.5222} &0.4763 &\quad &\textbf{0.5180} &0.4962\\
\quad &0.5 &\quad &\textbf{0.4951} &0.4826 &\quad &0.4974 &\textbf{0.5258}\\
\quad &0.75 &\quad &0.4240 &\textbf{0.4774} &\quad &\textbf{0.4995} &0.4955\\
\quad &1.0 &\quad &0.3923 &\textbf{0.4825} &\quad &0.4672 &\textbf{0.5323}\\

\hline
\multirow{5}{*}{CIFAR-10} &0.0 &\multirow{5}{*}{\textcolor{mygray}{0.1347}} &\textbf{0.1305} &0.1275 &\multirow{5}{*}{\textcolor{mygray}{0.1437}} &\textbf{0.1439} &0.1283\\
\quad &0.25 &\quad &\textbf{0.1366} &0.1275 &\quad &\textbf{0.1491} &0.1376\\
\quad &0.5 &\quad &0.1252 &\textbf{0.1307} &\quad &0.1316 &\textbf{0.1411}\\
\quad &0.75 &\quad &0.1303 &\textbf{0.1360} &\quad &0.1197 &\textbf{0.1464}\\
\quad &1.0 &\quad &0.1147 &\textbf{0.1341} &\quad &0.1237 &\textbf{0.1494}\\

\hline
\multirow{5}{*}{STL-10} &0.0 &\multirow{5}{*}{\textcolor{mygray}{0.1550}} &0.1390 &\textbf{0.1533} &\multirow{5}{*}{\textcolor{mygray}{0.1602}} &\textbf{0.1514} &0.1505\\
\quad &0.25 &\quad &0.1361 &\textbf{0.1448} &\quad &0.1479 &\textbf{0.1527}\\
\quad &0.5 &\quad &\textbf{0.1505} &0.1377 &\quad &0.1112 &\textbf{0.1620}\\
\quad &0.75 &\quad &0.1256 &\textbf{0.1513} &\quad &0.1001 &\textbf{0.1603}\\
\quad &1.0 &\quad &0.1328 &\textbf{0.1527} &\quad &0.1351 &\textbf{0.1553}\\

\hline
\multirow{5}{*}{Pendigits} &0.0 &\multirow{5}{*}{\textcolor{mygray}{0.6523}} &0.7079 &\textbf{0.7444} &\multirow{5}{*}{\textcolor{mygray}{0.6521}} &0.6523 &\textbf{0.6611}\\
\quad &0.25 &\quad &0.6420 &\textbf{0.7304} &\quad &\textbf{0.6535} &0.6469\\
\quad &0.5 &\quad &\textbf{0.6285} &0.6149 &\quad &\textbf{0.6823} &0.6349\\
\quad &0.75 &\quad &0.4493 &\textbf{0.6228} &\quad &0.6323 &\textbf{0.7364}\\
\quad &1.0 &\quad &\textbf{0.5222} &0.5063 &\quad &\textbf{0.6772} &0.5062\\

\hline
count &- &- &9 &16 &- &10 &15 \\

\hline\hline
\end{tabular}\label{Kappa}}
\end{table}

\subsection{Effectiveness analysis of SDA-FC}
To evaluate the effectiveness of SDA-FC, we employ two state-of-the-art federated clustering methods as baselines: k-FED \cite{dennis2021heterogeneity} and federated fuzzy c-means (FFCM) \cite{stallmann2022towards}. Additionally, to contextualize the performance of federated clustering relative to centralized clustering, we report numerical results for K-means (KM) and fuzzy c-means (FCM) in centralized scenarios, denoted as \textit{KM\_central} and \textit{FCM\_central}, respectively. In all experiments, the default fuzzy degree in FCM-based methods is set to 1.1.

\begin{figure}[!t]
\centering
\includegraphics[height = 4cm, width = 6cm]{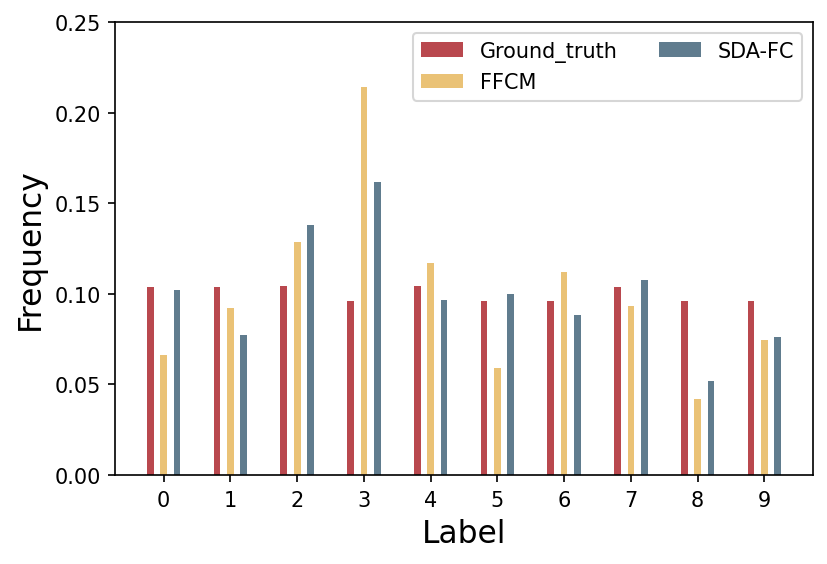}
\caption{The label distributions of different partitions on the dataset Pendigits with the Non-IID level p = 0.}
\label{case_kappa}
\end{figure}

The clustering performance based on NMI \cite{strehl2002cluster} and kappa \cite{liu2019evaluation} is shown in Table \ref{NMI} and Table \ref{Kappa}.  One can observe that: 1) For the KM-based methods, both metrics consistently indicate the superiority of our proposed method over k-FED in terms of both effectiveness and robustness. For the FCM-based methods, the two metrics give different ranks. Although NMI is the most commonly used metric, recent researchs highlight its limitations, rendering Kappa a more reliable alternative \cite{liu2019evaluation,yan2022selective}. Here, we also contribute some new cases to show that NMI fails to accurately reflect the performance of clustering results. As shown in Fig. \ref{case_kappa}, SDA-FC provides clustering results more aligned with the ground truth, yet NMI ranks it lower (0.6806 for FFCM and 0.6654 for SDA-FC), which is perplexing. In contrast, Kappa produces a more reasonable ranking (0.6523 for FFCM and 0.6611 for SDA-FC). Moreover, on the CIFAR-10 dataset, we establish an oracle baseline with ground-truth centroids (computed using ground-truth labels) in a centralized scenario, which assigns clusters based on cosine distance from local data to the centroids. It is intuitive that KM-based or FCM-based methods cannot surpass the oracle baseline. However, NMI ranks them inversely (0.1009 for the oracle baseline, 0.1051 for FFCM with a fuzzy degree of 2 when p = 0.75, and 0.1059 for FFCM with a fuzzy degree of 2 when p = 1), while Kappa provides more reasonable values (0.1992 for the oracle baseline, 0.1634 for FFCM with a fuzzy degree of 2 when p = 0.75, and 0.1854 for FFCM with a fuzzy degree of 2 when p = 1). Hence, for FCM-based methods, our proposed approach excels in terms of both effectiveness and robustness. 2) There exists a discernable gap between federated clustering and centralized clustering, but the SDA-FC framework mitigates this gap.

\begin{figure*}[!ht]
\centering
\includegraphics[height = 9cm, width = 18cm]{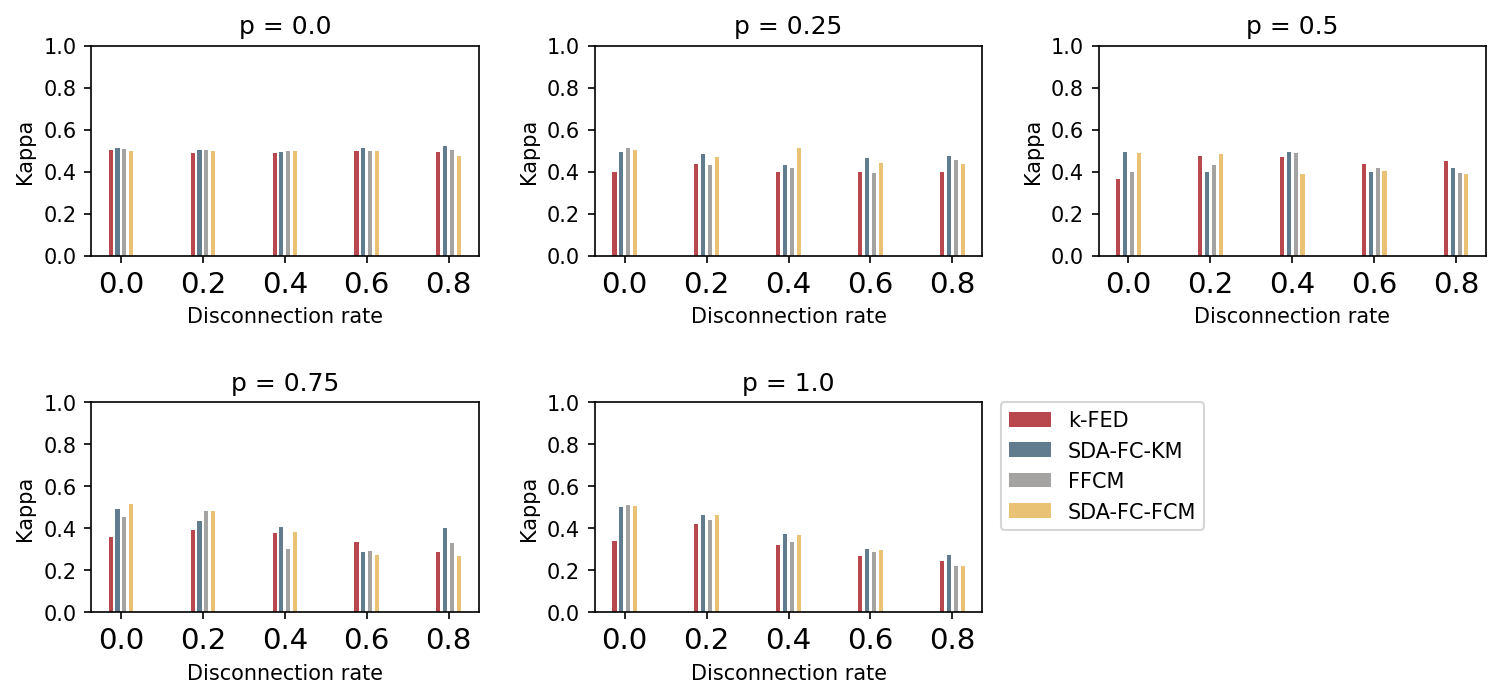}
\caption{The relations between the clustering performance and the device failures.}
\label{case_ft}
\end{figure*}

\begin{figure}[!t]
\centering
\subfigure[MNIST]{
\includegraphics[width=1.5in]{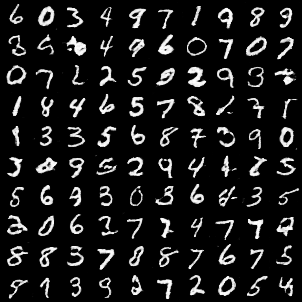}}
\quad
\subfigure[Fashion-MNIST]{
\includegraphics[width=1.5in]{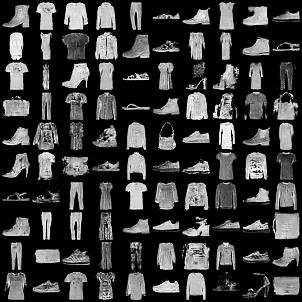}}
\quad

\subfigure[CIFAR-10]{
\includegraphics[width=1.5in]{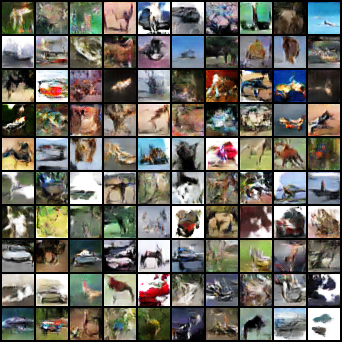}}
\quad
\subfigure[STL-10]{
\includegraphics[width=1.5in]{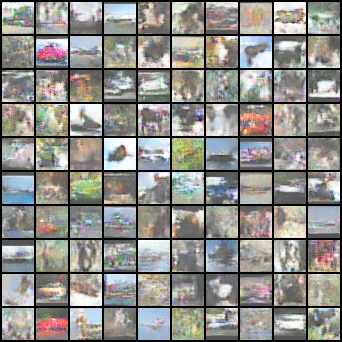}}
\caption{Visualization of the global synthetic datasets with the Non-IID level $p = 0.5$.}
\label{global_fake_samples}
\end{figure}

\begin{figure}[!t]
\centering
\subfigure[MNIST]{
\includegraphics[width=1.5in]{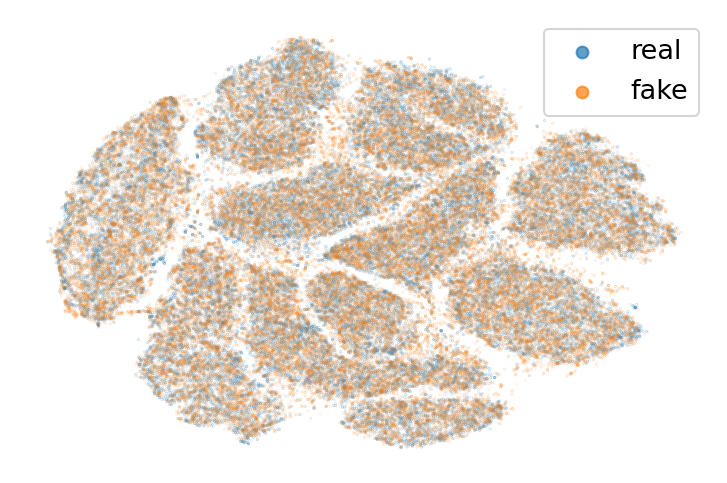}}
\quad
\subfigure[Fashion-MNIST]{
\includegraphics[width=1.5in]{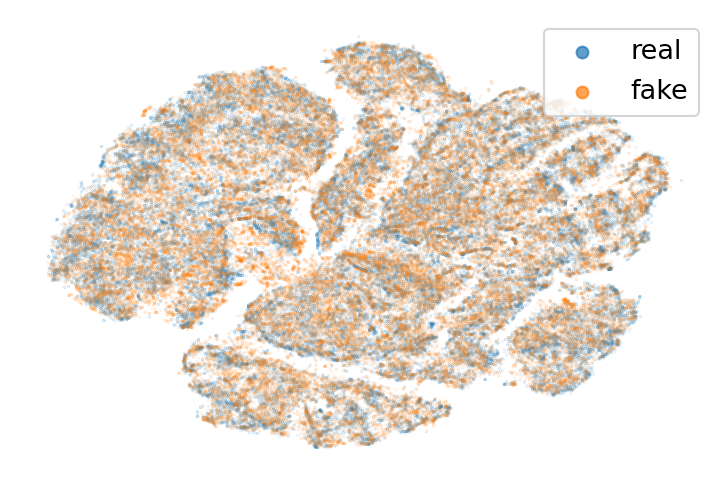}}
\quad

\subfigure[CIFAR-10]{
\includegraphics[width=1.5in]{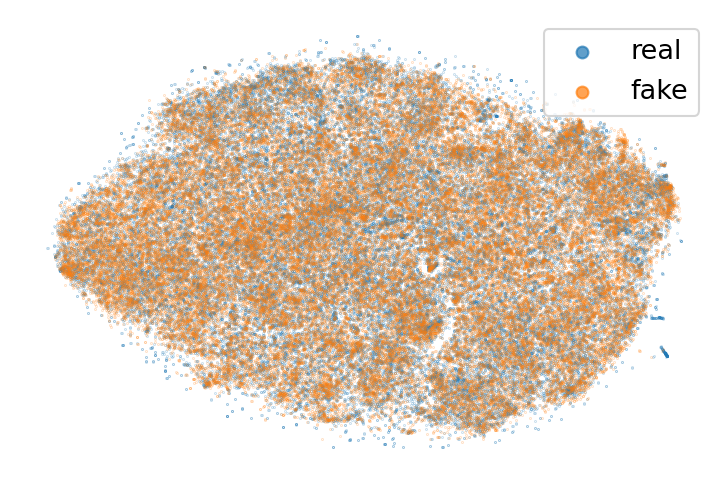}}
\quad
\subfigure[STL-10]{
\includegraphics[width=1.5in]{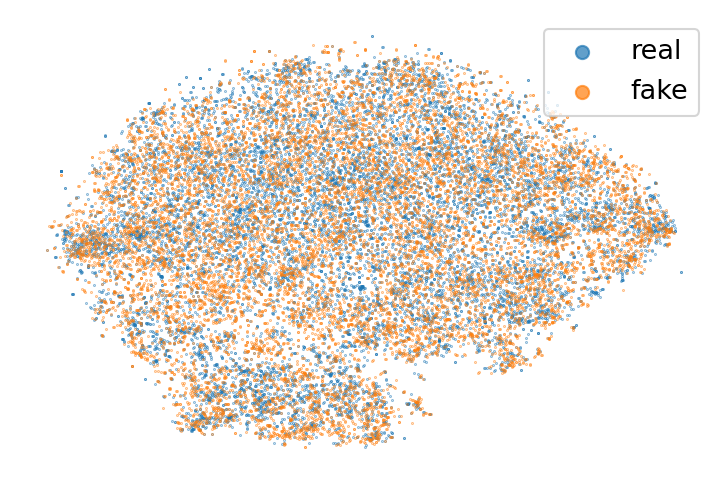}}
\caption{t-SNE visualization of the global synthetic datasets with the Non-IID level $p = 0.5$.}
\label{global_fake_distributions}
\end{figure}

\subsection{Effectiveness analysis of the global synthetic dataset}
While we have demonstrated the effectiveness of the SDA-FC framework in conjunction with KM or FCM, we expect that SDA-FC can serve as a versatile framework for addressing more intricate challenges in federated clustering, including its integration with deep clustering methods. The crux lies in ensuring that the global synthetic dataset generated by SDA-FC is a good approximation of the real one.

For each image dataset, we generate a global synthetic dataset of the same size as the real dataset. We showcase part of the generated images randomly in Fig. \ref{global_fake_samples} and visualize the global data distribution using t-SNE \cite{van2008visualizing} in Fig. \ref{global_fake_distributions}. One can see that: 1) Despite instances where objects in generated images are unrecognizable, SDA-FC effectively captures the fundamental data characteristics. 2) The global synthetic datasets and the global real ones exhibit substantial overlap, affirming that the former serves as a good approximation of the latter. 3) The cluster structure is more evident in grayscale image datasets compared to color image datasets. This elucidates the superior clustering performance observed in grayscale datasets, as outlined in Table \ref{Kappa}.

\subsection{Sensitivity analysis of clustering performance to device failures}
In practice, certain client devices may experience disconnections from the server due to factors such as wireless network fluctuations or energy constraints. Consequently, specific data characteristics from the disconnected devices may be lost, potentially affecting clustering performance. This means that the sensitivity analysis of clustering performance to device failures is worth investigating.

To handle such problem, following \cite{li2020federated}, we simply ignore the failed devices and continue training with the remaining ones. The rate of disconnection among all devices is denoted as the \textbf{disconnection rate}. Our experiments simulate various disconnected scenarios on MNIST by scaling the disconnection rate and run four federated clustering methods on them. As shown in Fig. \ref{case_ft}, one can see that: 1) The proposed methods exhibit superior effectiveness and robustness in all considered scenarios. 2) The sensitivity of clustering performance to device failures is positively correlated with the non-IID level $p$. Larger $p$ values result in reduced mutual substitutability of data characteristics across clients (one extreme case is that the data across the clients is completely non-IID when $p = 1$), exacerbating the adverse impact of device failures on clustering performance.

In summary, our findings reveal that: 1) The proposed methods outperform baselines in terms of both effectiveness and robustness. 2) Kappa emerges as a more reliable metric compared to NMI. 3) The global synthetic dataset generated by local GANs serves as a commendable approximation of the global real dataset. 4) Our proposed method demonstrates heightened resilience to device failures. 5) The sensitivity of clustering performance to device failures intensifies with higher heterogeneity levels, as characterized by $p$.

\section{Conclusion}
\label{sect6}
In this study, we have introduced SDA-FC, a simple yet effective federated clustering framework with GAN-based data synthesis. SDA-FC demonstrates outstanding performance in terms of effectiveness and robustness, surpassing existing state-of-the-art baselines. Notably, it runs with minimal communication overhead, requiring just one round and supports asynchronous execution. Furthermore, by leveraging a global synthetic dataset, our proposed methods exhibit superior performance in addressing non-IID data challenges and device failures without sharing private data.

Nonetheless, it is essential to acknowledge that SDA-FC, when paired with shallow clustering methods, may face limitations when dealing with complex data types, such as color images. Future research endeavors could explore the integration of SDA-FC with deep clustering methods to tackle these more intricate scenarios. Lastly, our findings underline the importance of considering appropriate evaluation metrics. Recent studies have cast doubts on the reliability of NMI values as indicators of clustering performance. We advocate for the adoption of the Kappa metric as a more robust and accurate alternative for assessing clustering results.

\bibliographystyle{named}
\bibliography{references}

\end{document}